\documentclass[10pt,twocolumn,letterpaper]{article}

\usepackage{cvpr}              % To produce the CAMERA-READY version
\usepackage{xcolor}
\usepackage{epsfig}
\usepackage{subcaption}
\usepackage{calc}
\usepackage{amssymb}
\usepackage{amstext}
\usepackage{amsmath}
\usepackage{amsthm}
\usepackage{multicol}
\usepackage{multirow}
\usepackage{algorithm2e}
\usepackage[bottom]{footmisc}

\usepackage{booktabs} 
\usepackage{graphicx}

\definecolor{cvprblue}{rgb}{0.21,0.49,0.74}
\usepackage[pagebackref,breaklinks,colorlinks,allcolors=cvprblue]{hyperref}

\def\paperID{*****} % *** Enter the Paper ID here
\def\confName{CVPR}
\def\confYear{2025}

\title{Segment-Level Road Obstacle Detection \\ Using Visual Foundation Model Priors and Likelihood Ratios}

\author{
Youssef Shoeb$^{1,3}$ \quad Nazir Nayal$^2$ \quad Azarm Nowzad$^{1}$ \\ \quad Fatma Güney$^2$ \quad Hanno Gottschalk$^3$\\
$^1$ Continental AG, Germany \\ \quad$^2$ Koç University, Türkiye \\ $^3$ Technische Universität Berlin, Germany \quad \\
{\tt\small youssef.shoeb@continental.com}
}

\begin{document}
\maketitle
\begin{abstract}
Detecting road obstacles is essential for autonomous vehicles to safely navigate dynamic and complex traffic environments.
Current road obstacle detection methods typically assign a score to each pixel and apply a threshold to generate final predictions.
However, selecting an appropriate threshold is challenging, and the per-pixel classification approach often leads to fragmented predictions with numerous false positives.
In this work, we propose a novel method that leverages segment-level features from visual foundation models and likelihood ratios to predict road obstacles directly.
By focusing on segments rather than individual pixels, our approach enhances detection accuracy, reduces false positives, and offers increased robustness to scene variability.
We benchmark our approach against existing methods on the RoadObstacle and LostAndFound datasets, achieving state-of-the-art performance without needing a predefined threshold.
\end{abstract}    
\section{Introduction}
\label{sec:intro}

Detecting road obstacles is critical for ensuring the safe manoeuvring of automated vehicles. Deep Neural Networks (DNNs) have demonstrated impressive performance on various perception tasks in automated driving, such as traffic sign recognition, road segmentation, and object detection. However, DNN-based approaches tend to perform poorly on detecting objects not encountered in their training data~\cite{nguyen2015deep}. This presents a significant safety concern, as including all potential road obstacles in the training data is impractical and can lead to potentially hazardous situations on the road if a road obstacle is missed.

\begin{figure*}[t!]
    \centering
    \includegraphics[width=1.\textwidth]{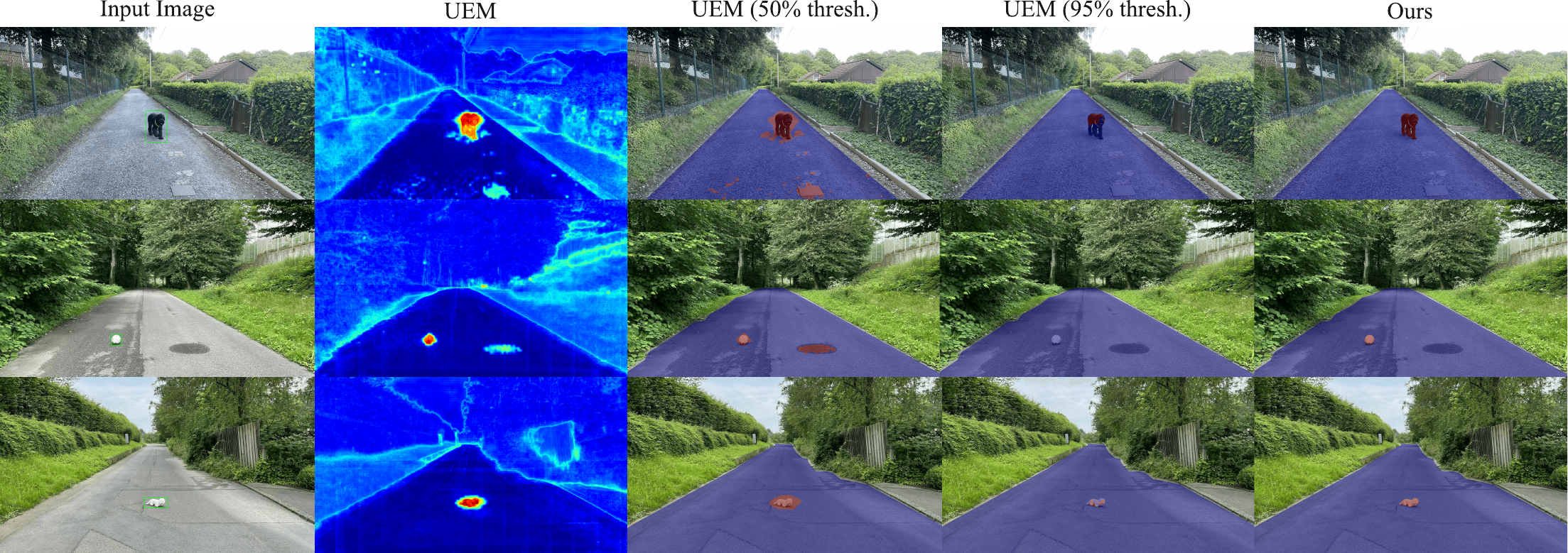}
    \vspace{5pt}
    \captionof{figure}{\textbf{Road Obstacle Segmentation Overview}. 
    From the \textbf{input image} (anomaly highlighted with a green box), current SOTA per-pixel methods (e.g., \textbf{UEM~\cite{nayal2024likelihoodratiobasedapproachsegmenting}}) produce high anomaly scores for unknown objects (column two), but when a threshold is applied the output is fragmented with multiple false positives or with false negatives if the threshold is set too low or too high (column three and four). \textbf{SAM} produces high-quality segment masks for all image segments but lacks semantic information. Our method (column five) uses the object priors used in SAM to learn the semantic distribution of the segments and detect the road obstacle segments based on the likelihood ratios. 
    }
    \vspace{-12pt}
    \label{fig:intro}
\end{figure*}

For the task of semantic segmentation, learned features are densely mapped to a pre-defined set of classes by a pixel-level classifier, allowing for accurately detecting and localizing every object in the image. Since training a segmentation model for all possible road obstacles is infeasible, road obstacle detection has been commonly addressed as an out-of-distribution (OoD) detection task. Previous methods for OoD detection in semantic segment networks have primarily focused on per-pixel reasoning~\cite{9578249,pebal_2022_eccv,nayal2024likelihoodratiobasedapproachsegmenting}, where each pixel is processed independently without considering the objectness of the segment to which the pixel belongs. More recent work ~\cite{ackermann2023maskomaly,nayal2023ICCV} attempted to resolve this issue by using mask-based semantic segmentation networks. These methods have shown promising results in preserving the objectness of objects in the training set and accurately identifying anomalous pixels. However, they still struggle to segment the OoD object as a whole effectively. We argue that while these methods are trained to detect instances of the in-distribution classes, the OoD objects are detected only as the residual pixels not detected by any of the masks. As an alternative in this work, we tackle the road obstacle detection task in semantic segmentation networks using segment-level reasoning and features obtained from visual foundation models.

Previous methods provide pixel-level scoring for discriminating between in-distribution classes and OoD objects. The separability of this score is then used as the main evaluation metric for per-pixel metrics. The standard per-pixel metrics (\textit{i.e.,} Average Precision, and FPR95) allow for evaluating performance in situations with a significant imbalance among classes, which is often the case for road obstacles. However, they tend to be biased towards larger obstacles, which is suboptimal as road obstacles can vary substantially in size, and each is equally important to detect. Component-level metrics are alternative evaluation metrics that serve as an indirect measure for object-level segmentation, assessing the overlap between the predicted and actual regions of anomalies. From a practitioner's perspective, these are the more interesting metrics to consider since most downstream tasks would need detections and not confidence scores.
Identifying the optimal threshold is often a challenging and complex task. Setting the threshold too low or too high can lead to either multiple false positives or missing detections (see \Cref{fig:intro}).

A common approach to threshold selection is to analyze the pixel-level precision-recall curve and select the value that maximizes the F1 score on a per-pixel basis. However, this approach requires a dedicated validation dataset and doesn't always result in optimal segment-level performance~\cite{chan2021segmentmeifyoucan}. Furthermore, even with the optimal threshold, the output masks produced by per-pixel detection methods often need further refinement, as some pixels may have inaccurate anomaly scores, resulting in fragmented or discontinuous masks.

In this work, we utilize the strong object priors in visual foundation models and present a method for road obstacle detection using segment-level features derived from the Segment Anything Model (SAM)~\cite{Kirillov_2023_ICCV}. Our approach generates more coherent and integrated segment-level predictions by focusing on segment features rather than individual pixels, addressing the inherent limitations of per-pixel predictions that often result in fragmented predictions. The final predictions are based on the likelihood ratio of two learned distributions: free-space and object segments. This allows us to mitigate the challenges associated with manual threshold selection and improves the overall robustness of the detection process.  

In summary, we summarize our contributions as follows: 
\begin{itemize}
    \item We introduce a novel road obstacle segmentation approach that leverages segment-level features from visual foundational models, moving beyond the pixel-level evaluation employed by existing methods.
    
    \item Our method utilizes a likelihood ratio between learned distributions, eliminating the need for manual threshold selection.
    
    \item We evaluate different approaches for learning the free-space and obstacle segment distributions and show that a non-parametric approach for approximating gives the best results.
    
    \item We demonstrate the effectiveness of our method in generalizing to unseen road obstacles and compare it to previous approaches on two benchmarks, outperforming all other methods on component-level metrics for both benchmarks.   
\end{itemize}

\section{Related Work}

\subsection{Road Obstacle Detection}
Previous approaches for road obstacle detection relied on multiple sensor modality setups to detect road obstacles. \cite{Williamson1998DETECTIONOS} used trinocular stereo vision and performed two types of stereo matching to determine whether a pixel belongs to a vertical or horizontal surface. \cite{7353537} used statistical hypothesis tests on local geometric features captured from a stereo vision system to detect obstacles. However, multi-camera systems present additional challenges, such as requiring exact calibration to perform image wrapping and computing the disparity between frames. In practice, vehicle vibration can complicate the calibration process since different cameras can move independently.

Other approaches required special types of sensors like Light detection and ranging (LiDAR) or radio detection and ranging (RADAR).  \cite{Tokudome2017DevelopmentOR} used LiDAR sensors to measure the reflection intensity of objects and detect road users. \cite{10160592} used RADAR signals for obstacle and free space detection. While utilizing special sensor modalities like LiDAR or RADAR signals could benefit obstacle detection, these specialized sensors are not always available in all vehicle perception systems due to costs and hardware limitations. 

In this work, we focus only on methods that operate on single-frame images captured by standard in-vehicle cameras as a promising alternative.

\subsection{Road Obstacle Segmentation}
The common approach for road obstacle segmentation relies on a robust closed-world segmentation model. This model is trained to detect a set of predefined classes and to quantify an OoD score for each pixel that may belong to a different class. The per-pixel OoD score can be interpreted as a form of predictive uncertainty on the given training set. Earlier approaches modeled the uncertainty through maximum softmax probabilities~\cite{Hendrycks2017ICLR}, ensembles \cite{Lakshminarayanan2017NeurIPS}, Bayesian approximation~\cite{Mukhoti2018ARXIV}, 
 or Monte Carlo dropout~\cite{Gal2016ICML}. However, the posterior probabilities produced by a model trained in a closed-word setting may not always be well-calibrated, often resulting in overly confident predictions for unseen categories~\cite{Nguyen2015CVPR,Guo2017ICML,Minderer2021NeurIPS,Jiang2018NeurIPS}. In this work, we utilize the strong object priors that visual foundation models learn during their training and utilize this to predict road obstacles directly, without having a closed-world segmentation model. 

\cite{hendrycks2019oe} introduced outlier exposure as a strategy for enhancing the performance of OoD detection. Outlier exposure leverages a proxy dataset composed of outliers to discover signals and learn heuristics for OoD samples. 
\cite{nayal2023ICCV} used a proxy dataset to train the model to produce low logit scores on unknown objects. We follow a similar approach in our work, relying on a proxy dataset, but we explicitly try to model the proxy distribution of potential road obstacles and use this to differentiate between free-space and obstacle segments. 

\subsection{Nearest-Neighbour OoD Detection}
Retrieval-based methods have been explored for anomaly detection \cite{reiss2021panda,retrieval_industry,Zou_22}, relying on large samples of in-distribution datasets to identify anomalies as deviations from the expected data patterns.  \cite{sun2022knnood} highlighted the potential of using \textit{k-nearest-neighbors} (KNN) for OoD detection in deep neural networks. They used KNNs to calculate the distance between the embedding of each test image and the training set, then applied a threshold-based criterion to decide whether an input is OoD.
However, their exploration was limited to an image recognition context, which is characterized by single-instance, object-centric images. \cite{10350977} extended the application of KNN to transformer-based representations, achieving state-of-the-art performance on common driving-focused anomaly detection benchmarks. One limitation of their approach was its low resolution, which limits its utility and applicability. Our work adopts a similar strategy and uses KNNs to learn feature representations from transformer-based models. However, we learn two explicit distributions: one for free-space and another for road obstacles. This approach enhances the model's ability to distinguish between road obstacles and road segments more precisely.

\subsection{Open-World Segmentation}

Open-world segmentation seeks to segment all objects in the image, even those not in their training dataset. Recent advancements in large-scale, text-guided training for classification \cite{pmlr-v139-jia21b,pmlr-v139-radford21a} have inspired several studies to adapt and extend these methodologies to the domain of open-world segmentation \cite{rao2021denseclip,ding2023maskclip,9879676}. However, a limitation of these approaches is their reliance on text prompts to segment objects.  SEEM \cite{zou2023segment} and  Segment Anything Model (SAM) \cite{Kirillov_2023_ICCV} build upon previous work and allow for various types of prompts. In our approach, we leverage SAM to generate and represent regions. Similar to our work, \cite{nekrasov2023ugains} also used SAM for road obstacle detection, but they relied on an OoD segmentation model to identify unknown regions and use this to prompt SAM. In our approach, we directly use SAM features to detect road obstacles which streamlines the process and potentially reduces reliance on secondary models. 

\begin{figure*}[tb]
  \centering
  \includegraphics[width=\linewidth]{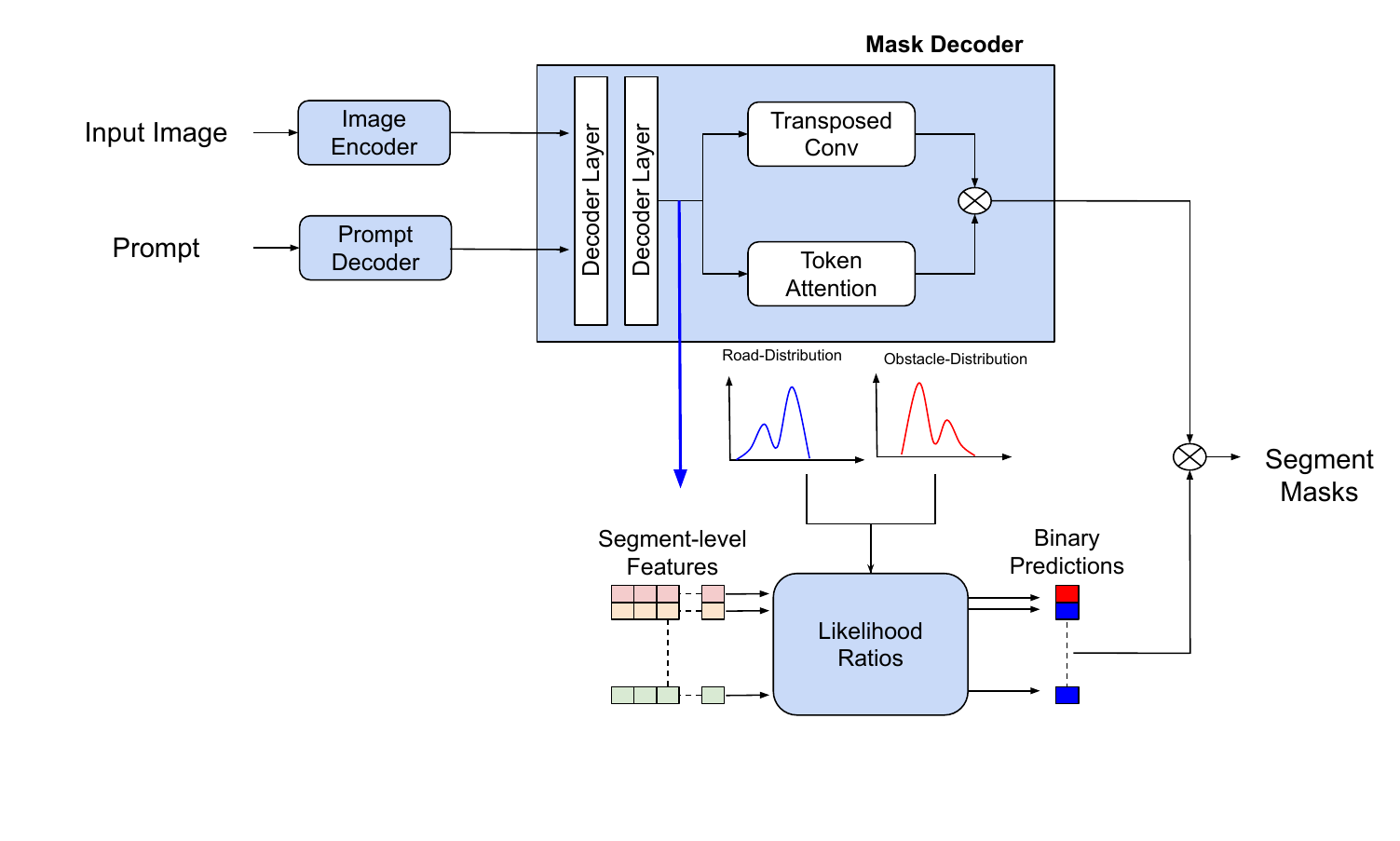}
  \caption{\textbf{Approach For Segment-Level Road Obstacle Detection:} Our approach for road obstacle detection uses visual foundation models like SAM \cite{Kirillov_2023_ICCV} to generate segment-level masks. The segment-level feature representations are obtained from the transformer decoder layer, which processes the image and prompts embeddings. During inference, we generate masks for the entire image using a grid of point prompts over the image and filter low-quality and duplicate masks outside the region of interest. For each remaining mask, we compute the likelihood ratios of these learned representations to produce final predictions using two learned estimates trained to estimate free space and obstacles.
  }
  \label{fig:our_method}
\end{figure*}

\section{Method Description}

We present our method in this section (see \cref{fig:our_method}); we first give an overview of SAM and how we extract segment-level feature representations. Next, we introduce our task formulation for road obstacle detection, focusing on how we leverage likelihood ratios to differentiate between two learned distributions: one for free-space segments and another for proxy road obstacles. This approach allows us to detect road obstacles more robustly by directly operating on segment-level features, which mitigates the issues commonly encountered with per-pixel predictions.

\subsection{Preliminaries: Overview of SAM}

SAM was recently introduced as a foundational vision model for general image segmentation. It was trained on the large-scale SA-1B dataset, which contains over 1 billion masks from 11 million images. SAM's architecture comprises three main modules: 1) an image encoder for extracting image features, 2) a prompt encoder that encodes positional information from the input, and 3) a mask decoder that combines the image features and prompt tokens to generate final mask predictions. Experimental results show powerful zero-shot capabilities to segment a wide range of objects, parts, and visual structures across diverse scenarios. Therefore, an interesting question arises: \emph{can we utilize SAM's strong object priors for learning semantic features of objects and regions?}

However, efficiently extracting semantics from visual foundation models is a non-trivial challenge. While a simple solution might be to use feature embeddings directly from the image encoder, we argue that the prompt priors contained in the prompt tokens are critical for accurately segmenting object boundaries. Therefore, we extract segment-level representations from the intermediate layers of the mask decoder, specifically after the transformer decoder layers and convolution (blue arrow in \cref{fig:our_method}). Each segment-level representation is a vector of size 2048, encoding both the intersection over union prediction and mask positions. 

\subsection{Road Obstacle Detection Using Likelihood Ratios}

A simple approach for road obstacle detection is to learn a density model $p_{free}$ for free-space segments and predict an obstacle when the likelihood $p_{free}(x)$ of the input features $x$ for is low (\textit{i.e.,} there is little training data in the region around $x$ ). However, since we utilize neural networks that abstract information and produce a condensed representation for each input, utilizing only a single distribution for the task may lead to unreliable results. As shown by~\cite{nalisnick2018deep}, a density estimate learned on one dataset may assign higher scores to inputs from a completely different dataset (\textit{i.e.,} one distribution may sit inside of another distribution due to the feature extraction process). This behaviour suggests that using a single distribution may fail to distinguish between free-space and road obstacles effectively.         

We formulate the road obstacle detection task as a model selection task between two distributions representing the free-space segments $P_{free}$ and obstacles $P_{obstacle}$. Given a feature vector $x$, consider the null hypothesis $\mathcal{H}_{free}$ that $x$ was drawn from $P_{free}$, and an alternate hypothesis $\mathcal{H}_{obstacle}$ that $x$ was drawn from $P_{obstacle}$. By the Neyman-Pearson lemma~\cite{neyman1933ix}, when fixing type-I errors (false-positives), the test with the smallest type-II errors (false-negatives) is the likelihood ratio test.
\begin{equation}
    LR=\frac{p_{obstacle}(x)}{p_{free}(x)}
\end{equation}
\cite{zhang2022falsehoods} showed that the same conclusion holds for a Bayesian perspective. Directly predicting the final output from this ratio is a formulation of the \textit{Maximum Likelihood} (ML) decision rule~\cite{+1996} from decision theory.  
Since road obstacles are very rare in practice, this would mean that the ML would overestimate the likelihood of road obstacles. 
However, from a safety perspective, it would be more desirable to base the detections only on the most likely observed features rather than potentially biasing our decision based on prior knowledge.

While it is infeasible to estimate the true distribution 
$P_{obstacle}$ for every potential road obstacle—given the wide variability in obstacle types—we adopt the concept of \textit{outlier exposure}~\cite{Hendrycks2019ICLR} to approximate  $P_{obstacle}$ using a proxy dataset. This proxy dataset captures a diverse set of possible obstacles, allowing us to model the distribution without accounting for every scenario. Importantly, in our formulation, the obstacle distribution does not need to be perfectly precise. Rather, it only needs to exhibit greater similarity to the proxy dataset than to the distribution of free-space segments, making it sufficient for effective obstacle detection.

\subsection{Distribution Estimation Methods}
We build two \textit{reference feature} datasets for the free-space segments and out-distribution datasets. Both datasets are obtained from an internal dataset that contains labelled road obstacles captured from a real-world test vehicle in both urban and highway driving conditions.
The reference feature datasets are denoted as $\mathcal{R}^{free} \in \mathbb{R}^{N\times C}$ and $\mathcal{R}^{obstacle} \in \mathbb{R}^{M\times C}$ where $N$ and $M$ are the number of reference features and $C$ is the dimensionality of each reference features. For both datasets, the dimensionality $C$ is 2048, and the number of reference features is 10k.   
We evaluate three distinct approaches for estimating the distributions of these datasets: Gaussian Mixture Models (GMMs), which provide log-likelihood estimates; Normalizing Flows, which offer exact density estimates; and k-Nearest Neighbours (k-NN), which compute odds estimates based on neighborhood distances of the feature representations. We dive into the details of each method in this section.

\subsubsection{Gaussian Mixture Models}
Gaussian Mixture Models (GMMs) assume that the feature vectors in a dataset are generated from a mixture of several Gaussian distributions. Each Gaussian distribution in the mixture represents a different cluster, characterized by its own mean vector and covariance matrix. Formally, the probability density function of a GMM is expressed as:

\[
p(x \mid \lambda) = \sum_{i=1}^{K} \pi_i \mathcal{N}(x \mid \mu_i, \Sigma_i)
\]

where \(K\) is the number of Gaussian components, \(\pi_i\) represents the mixing coefficients (such that \(\sum_{i=1}^{K} \pi_i = 1\)), \(\mu_i\) is the mean, and \(\Sigma_i\) is the covariance matrix of the \(i\)-th component.

We fit two separate GMMs, on \(\mathcal{R}^{in}\) and on \(\mathcal{R}^{out}\), to model the in-distribution and out-distribution feature sets, respectively. The GMM parameters \(\lambda^{in} = \{ \pi^{in}_i, \mu^{in}_i, \Sigma^{in}_i \}\) are estimated from \(\mathcal{R}^{in}\), while the parameters \(\lambda^{out} = \{ \pi^{out}_i, \mu^{out}_i, \Sigma^{out}_i \}\) are estimated from \(\mathcal{R}^{out}\).

During inference, we compute the likelihood of each segment-level feature \(t\) under both the in-distribution and out-distribution models. This is done by calculating the likelihood of \(t\) under each GMM:

\[
p(t \mid \lambda^{in}) =  \sum_{i=1}^{K} \pi^{in}_i \mathcal{N}(t \mid \mu^{in}_i, \Sigma^{in}_i)
\]
\[
 p(t \mid \lambda^{out}) =  \sum_{i=1}^{K} \pi^{out}_i \mathcal{N}(t \mid \mu^{out}_i, \Sigma^{out}_i)
\]

We then compare the odds estimate of both distributions. If $p(t \mid \lambda^{in}) / p(t \mid \lambda^{out}) \geq 1$, we predict that the test sample \(t\) belongs to the in-distribution. 
Otherwise, we classify it as belonging to the out-distribution.

We use a GMM with \(K = 50\) components and a diagonal covariance matrix, which balances model complexity and accuracy in distinguishing between in-distribution and out-distribution samples. The parameters of the GMM are estimated using the Maximum Likelihood estimator for the observed data using the Expectation Maximization (EM) algorithm~\cite{dempster1977maximum} with initialization done using K-means clustering.

\subsubsection{Normalizing Flows}
Normalizing Flows (NF) are a generative model class that uses a sequence of invertible transformations to map a simple prior distribution (e.g., Gaussian) to a more complex distribution that fits the data. Each transformation in the flow is designed to be both invertible and differentiable, allowing for the computation of both the density and sampling~\cite{9089305}. We learn a flow for each distribution and use the density estimates to predict which distribution the sample belongs to.  

In our approach, we use Neural Spline Flows~\cite{NEURIPS2019_7ac71d43}, specifically coupling layers based on rational quadratic splines. The splines parameterize piecewise invertible functions, making mapping between simple and complex distributions possible. The transformations are conditioned on half of the input dimensions, learned via a four-layer MLP. 

The overall flow is constructed by a sequence of three blocks, where each block includes an actnorm layer, an Invertible 1x1 Convolution~\cite{pmlr-v97-hoogeboom19a}, and a Neural Spline Flow. The flow is designed to map the input data distribution to a standard Gaussian prior \( \mathcal{N}(0, 1)\). During inference, each segment-level feature \(t\) is passed through the model transforms, and the log probability score is computed for each model, approximating the free-space and road obstacle. If the log probability scores for the road obstacle model are larger than the log probability scores for the free-space model, then we predict the segment to be an obstacle.

\subsubsection{$K$-Nearest Neighbors} 
$k$-Nearest Neighbors relies on the computation of distances between feature representations as a measure of estimation. More formally, each segment-level feature $t$ of the image is extracted, and the distance to each of the features in $\mathcal{R}^{in}$ and $\mathcal{R}^{out}$ denoted as $\text{dist}(t, x^{in})$ and $\text{dist}(t, x^{out})$. We then find the $k$ samples with the highest cosine similarity values to $t$ in $\mathcal{R}^{in}$ and $\mathcal{R}^{out}$. Let $N_{\mathcal{R}^{in}}^k(t)$ and $N_{\mathcal{R}^{out}}^k(t)$ be the sets of the top-$k$ most similar samples to $t$ based on the cosine similarity values. The average cosine similarity between $t$ and the top-$k$ most similar samples in $\mathcal{R}$ are calculated as:

\begin{equation}
  \overline{\text{dist}}(t, N_{\mathcal{R}}^k(t)) = \frac{1}{k} \sum_{x_i \in N_{\mathcal{R}}^k(t)} \text{dist}(t, x_i)  
\end{equation}
If $   \frac{\overline{\text{dist}}(t, N_{\mathcal{R}^{obstacle}}^k(t))}{\overline{\text{dist}}(t, N_{\mathcal{R}^{free}}^k(t))} \geq 1 $
we predict that the test sample $t$ is more likely to be an obstacle. We use the cosine similarity as a distance metric and find $k$ equal to five to give the best results.  

\section{Experiments}

\begin{table*}[ht]
\setlength{\tabcolsep}{1.0em}
\centering
\resizebox{0.9\linewidth}{!}{
\begin{tabular}{lcccccc|ccccc}
\toprule
\multirow{2}{*}{Method} & \multicolumn{5}{c}{SMIYC-Obstacle} & & \multicolumn{5}{c}{SMIYC-LostAndFound}\\
\cmidrule(r){2-6} \cmidrule(r){8-12}
 & $AP\uparrow$ & $FPR_{95}\downarrow$ & $sIoU_{gt}\uparrow$ & $PPV\uparrow$ & $\overline{F}_1\uparrow$ & & $AP\uparrow$ & $FPR_{95}\downarrow$ & $sIoU_{gt}\uparrow$ & $PPV\uparrow$ & $\overline{F}_1\uparrow$ \\
\midrule
UEM~\cite{nayal2024likelihoodratiobasedapproachsegmenting} & \underline{94.40} & \underline{0.10} & 49.80 & 76.80 & 67.2 &  &   &   &   &   \\
UNO~\cite{delić2024outlierdetectionensemblinguncertainty} & 93.19 & 0.16 & \textbf{70.97} & 72.17 & \underline{77.65} & &  &  &  &  \\
RbA~\cite{nayal2023ICCV} & \textbf{95.12} & \textbf{0.08} & 54.34 & 59.08 & 57.44 &   &   &   &   &   \\
EAM~\cite{Grcic_2023_CVPR} & 92.87 & 0.52 & \underline{65.86} & 76.50 & 75.58 & &   &   &   &   &  \\
Mask2Anomaly~\cite{Rai_2023_ICCV} & 93.22 & 0.20 & 55.72 & 75.42 & 68.15 & &   &   &   &   &   \\
\midrule
NFlowJS~\cite{grcić2023dense} &   &   &   &   &   & & \textbf{89.28} & \textbf{0.65} & \textbf{54.63} & 59.74 & \underline{61.75} \\
PixOOD~\cite{Vojir_2024_ECCV} &   &   &   &   &   & & 85.07 & \underline{4.46} & 30.18 & \underline{78.47} & 44.41 \\
Road Inpainting~\cite{10334623} &   &   &   &   &   & & \underline{82.93} & 35.75 & 49.21 & 60.67 & 52.25 \\
SynBoost~\cite{9578249} &   &   &   &   &   & & 81.71 & 4.64 & 36.83 & 72.32 & 48.72 \\
DaCUP~\cite{Vojir_2023_WACV} &   &   &   &   &   &  & 81.37 & 7.36 & 38.34 & 67.29 & 51.14 \\
\bottomrule
LR (KNN) -ours- & 92.0 & 0.20 & 62.9 & \underline{81.9} & \textbf{78.4} & & 83.70 & - & \underline{49.70} & \textbf{95.90} & \textbf{72.60} \\
LR (GMM) -ours- & 91.90 & 0.20 & 59.5 & \textbf{84.2} & 76.90 & & 83.50 & - & 47.40 & 92.00 & 69.20 \\
LR (NF) -ours- & 77.10 & 33.70 & 46.50 & 77.5 & 62.70 & &76.90 & - & 44.70 & 73.50 & 61.00 \\

\end{tabular}
}

\vspace{5pt}
\caption{\textbf{Performance on SMIYC-Obstacle and SMIYC-LostAndFound Benchmarks}: We compare the performance of the top five state-of-the-art methods across both datasets. The best results are highlighted in \textbf{bold}, and the second best are \underline{underlined}. Our method achieves state-of-the-art performance on the $\overline{F}1$ score and $PPV$ in the component-level metrics. On pixel-level metrics, our method is not as competitive as state-of-the-art pixel segmentation networks due to obstacles that are missed and assigned as a road by default.}
\label{table:combined_SMIYC}
\end{table*}

\subsection{Experimental Setup}

\textbf{Metrics:} The standard metrics for pixel-level segmentation are Average Precision (AP) and False Positive Rate at a True Positive Rate of 95\% (FPR95). These are all threshold-independent metrics, which help evaluate the usability of a method irrespective of the chosen threshold. However, in practice, a threshold must always be selected for any downstream task that utilizes an OoD detector~\cite{maag2022two,shoeb2024have}.
Therefore, we focus on the three component-level metrics used in the SegmentMeIfYouCan (SMIYC) benchmark \cite{chan2021segmentmeifyoucan} (sIoU$_{gt}$, PPV, and mean F$_1$) as our main comparison metric. sIoU$_{gt}$ is the average intersection over union, measuring how well the prediction road obstacle overlaps with the ground truth. PPV is the average positive predictive value, and this asses how accurate the predicted road obstacles are (precision). Finally, mean F$_1$ is a balanced measure combining both sIoU and PPV. It is calculated at multiple thresholds and then averaged; this provides a single value reflecting both the ability to detect and the accuracy of these detections. 

\noindent \textbf{Datasets:} We evaluate our method on SMIYC-RoadObstacle, and SMIYC-LostAndFound. Both datasets represent realistic and hazardous obstacles on the road ahead that are critical to detect for an autonomous vehicle. SMIYC-RoadObstacle contains a total of 327 privately withheld images, which are used to evaluate different methods. SMIYC-LostAndFound is a filtered and refined- version of LostAndFound \cite{lost_and_found}.

\subsection{Comparison to State-of-the-Art}

We evaluate our proposed method with different distribution estimation methods, as shown in~\Cref{table:combined_SMIYC}~Table 1. Each approach is compared against the top five state-of-the-art methods on the SMIYC-Obstacle and SMIYC-LostAndFound datasets. Our results demonstrate that our method achieves competitive performance across all three distribution estimation techniques, with the non-parametric k-nearest neighbour approach yielding the best results on both datasets. The most significant improvement is observed on SMIYC-LostAndFound, where our method achieves a PPV of 95.9 and an $\overline{F}_1$ score of 72.60, outperforming the previous best method by 17.43 and 10.85, respectively.    

\begin{figure}[h]
    \centering
    \includegraphics[width=1\linewidth]{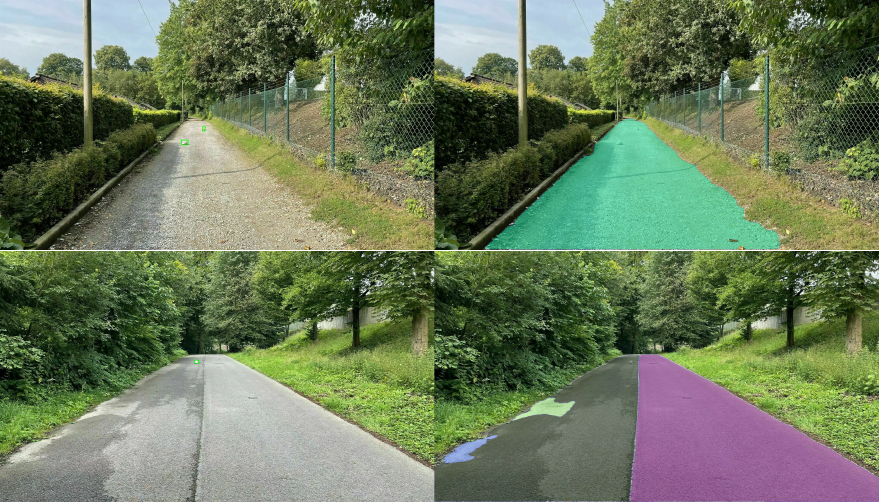}
    \caption{\textbf{Failure Case Examples:} The left column shows the input image with the road obstacles highlighted in green bounding boxes, and the right column shows scenarios where the masks generated by SAM miss detecting the road obstacle as a separate segment. }
    \label{fig:failcase}
\end{figure}

However, in pixel-level metrics such as average precision (AP) and false positive rate (FPR), our method is less competitive compared to state-of-the-art pixel segmentation networks like RbA and NFlowJS. This discrepancy can be attributed to the limitations of our model in detecting smaller or ambiguous obstacles, which are sometimes not detected as separate segments by SAM as seen in \Cref{fig:failcase}. Additionally, this causes the $FPR_{95}$ metric to fail on the SMIYC-LostAndFound dataset as more than 5\% of the obstacles are not detected as separate segments and are considered free space by default. Despite this, the superior performance in component-level metrics suggests that our method is highly effective at distinguishing between larger and more well-defined obstacles, which is critical for real-world applications of road obstacle detection.

\begin{figure*}[h]
    \centering
    \includegraphics[width=\textwidth]{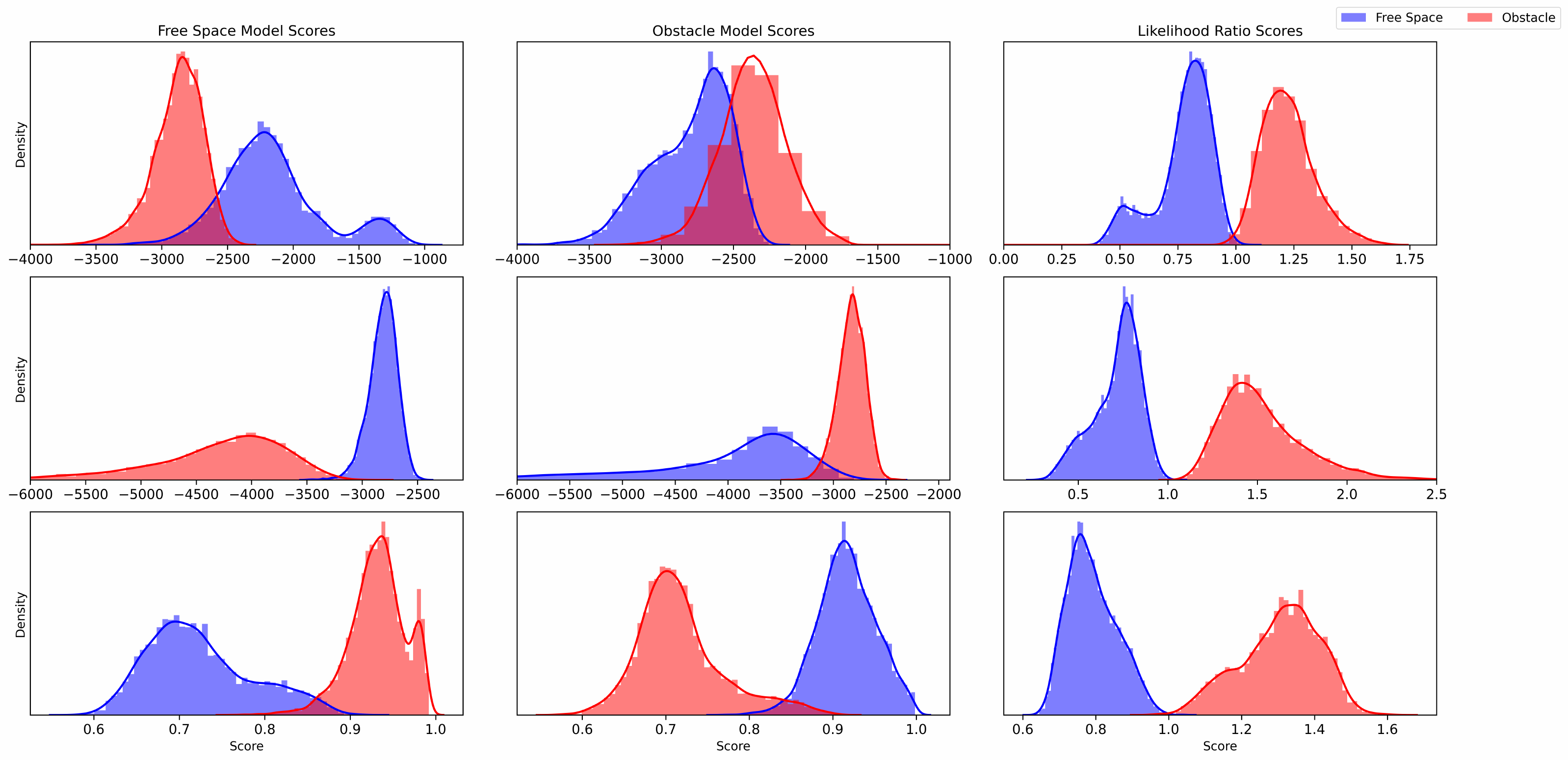}
    \caption{Comparison of Gaussian Mixture Models (first row) Normalizing Flows (second row), and K-nearest neighbours (third row) on the training set. The first column visualizes the learned distributions of the free-space model, the second visualizes the learned distributions of obstacles, and the third visualizes the likelihood ratio between both. The likelihood ratio provides better separation than any of the models separately at the threshold value 1. }
    \label{ablation}
\end{figure*}

\subsection{Ablations}
For each estimation method, we visualize the separation between the two learned distributions on the training data in \Cref{ablation}. The first row visualizes the likelihood of sampling from each GMM, the second row visualizes the density estimates for each normalizing flow model, and the final row visualizes one minus the average distance to the 5 nearest neighbours. For all three methods, utilizing the likelihood ratio between the two distributions provides a better separation than utilizing any one model on its own. We also find that the normalizing flow models provide the best separation of the training data. However, it does not generalize as well as the GMM or the k-nearest neighbours. 
\section{Conclusion \& Outlook}

In this paper, we propose a novel approach to address the road obstacle segmentation problem at the segment level. Our method leverages strong object priors from visual foundational models to generate segment-level features. We estimate the probability distributions for both \textit{free-space} and \textit{obstacle} segments and utilize the likelihood ratios as a binary classifier to detect road obstacles. We evaluate several methods for estimating these distributions, including GMMs, normalizing flows, and k-nearest neighbours, finding that k-nearest neighbours produce the best results. Our approach achieves state-of-the-art performance on standard benchmarks (SegmentMeIfYouCan and LostAndFound) in terms of component-level metrics without requiring a predefined threshold.  

\textbf{Limitations and Future Work:} Despite the strong performance in component-level metrics, our approach still suffers from a relatively high false-negative rate due to small objects not being detected as separate segments by SAM. This is due to the prompting strategy we deploy; the equally spaced grid may miss small objects that lie between the point prompts. Restricting the prompts to only regions where the model is uncertain could potentially resolve this issue, but then the segments that the model incorrectly classifies as the road will also be missed from the road obstacle detection module. Additionally, the efficacy of the best-performing method ($k$-nearest neighbours) relies on the number and quality of reference features selected. In this work, we utilized all available reference features without examining the impact of selecting different subsets of these features. Future work would investigate the application of more sophisticated techniques, such as core-set approaches \cite{Tereshchenko2024}, to choose the set of reference features selectively. By optimizing the selection process, we anticipated that the inference time of our models could be significantly enhanced without loss in performance.

\section*{Acknowledgements}
The research leading to these results is funded by the German Federal Ministry for Economic Affairs and Climate Action within the project “just better DATA". N.N. is funded by the KUIS AI Center, F.G. by the European Union (ERC, ENSURE, 101116486). Views and opinions expressed are, however, those of the author(s) only and do not necessarily reflect those of the European Union or the European Research Council. Neither the European Union nor the granting authority can be held responsible for them.
{
    \small
    \bibliographystyle{ieeenat_fullname}
    \bibliography{main}

\begin{thebibliography}{53}
\providecommand{\natexlab}[1]{#1}
\providecommand{\url}[1]{\texttt{#1}}
\expandafter\ifx\csname urlstyle\endcsname\relax
  \providecommand{\doi}[1]{doi: #1}\else
  \providecommand{\doi}{doi: \begingroup \urlstyle{rm}\Url}\fi

\bibitem[Ackermann et~al.(2023)Ackermann, Sakaridis, and Yu]{ackermann2023maskomaly}
Jan Ackermann, Christos Sakaridis, and Fisher Yu.
\newblock Maskomaly: Zero-shot mask anomaly segmentation.
\newblock In \emph{The British Machine Vision Conference (BMVC)}, 2023.

\bibitem[Chan et~al.(2021)Chan, Lis, Uhlemeyer, Blum, Honari, Siegwart, Fua, Salzmann, and Rottmann]{chan2021segmentmeifyoucan}
Robin Chan, Krzysztof Lis, Svenja Uhlemeyer, Hermann Blum, Sina Honari, Roland Siegwart, Pascal Fua, Mathieu Salzmann, and Matthias Rottmann.
\newblock Segmentmeifyoucan: A benchmark for anomaly segmentation.
\newblock In \emph{Thirty-fifth Conference on Neural Information Processing Systems Datasets and Benchmarks Track}, 2021.

\bibitem[Delić et~al.(2024)Delić, Grcić, and Šegvić]{delić2024outlierdetectionensemblinguncertainty}
Anja Delić, Matej Grcić, and Siniša Šegvić.
\newblock Outlier detection by ensembling uncertainty with negative objectness, 2024.

\bibitem[Dempster et~al.(1977)Dempster, Laird, and Rubin]{dempster1977maximum}
Arthur~P Dempster, Nan~M Laird, and Donald~B Rubin.
\newblock Maximum likelihood from incomplete data via the em algorithm.
\newblock \emph{Journal of the royal statistical society: series B (methodological)}, 39\penalty0 (1):\penalty0 1--22, 1977.

\bibitem[Di~Biase et~al.(2021)Di~Biase, Blum, Siegwart, and Cadena]{9578249}
Giancarlo Di~Biase, Hermann Blum, Roland Siegwart, and Cesar Cadena.
\newblock Pixel-wise anomaly detection in complex driving scenes.
\newblock In \emph{2021 IEEE/CVF Conference on Computer Vision and Pattern Recognition (CVPR)}, pages 16913--16922, 2021.

\bibitem[Durkan et~al.(2019)Durkan, Bekasov, Murray, and Papamakarios]{NEURIPS2019_7ac71d43}
Conor Durkan, Artur Bekasov, Iain Murray, and George Papamakarios.
\newblock Neural spline flows.
\newblock In \emph{Advances in Neural Information Processing Systems}. Curran Associates, Inc., 2019.

\bibitem[Fahrmeir et~al.(1996)Fahrmeir, Hamerle, and Tutz]{+1996}
Ludwig Fahrmeir, Alfred Hamerle, and Gerhard Tutz, editors.
\newblock \emph{Multivariate statistische Verfahren}.
\newblock De Gruyter, Berlin, Boston, 1996.

\bibitem[Gal and Ghahramani(2016)]{Gal2016ICML}
Yarin Gal and Zoubin Ghahramani.
\newblock Dropout as a bayesian approximation: Representing model uncertainty in deep learning.
\newblock 2016.

\bibitem[Galesso et~al.(2023)Galesso, Argus, and Brox]{10350977}
S. Galesso, M. Argus, and T. Brox.
\newblock Far away in the deep space: Dense nearest-neighbor-based out-of-distribution detection.
\newblock In \emph{2023 IEEE/CVF International Conference on Computer Vision Workshops (ICCVW)}, pages 4479--4489, Los Alamitos, CA, USA, 2023. IEEE Computer Society.

\bibitem[Grci\'c et~al.(2023)Grci\'c, \v{S}ari\'c, and \v{S}egvi\'c]{Grcic_2023_CVPR}
Matej Grci\'c, Josip \v{S}ari\'c, and Sini\v{s}a \v{S}egvi\'c.
\newblock On advantages of mask-level recognition for outlier-aware segmentation.
\newblock In \emph{CVPR Workshops}, 2023.

\bibitem[Grcić et~al.(2023)Grcić, Bevandić, Kalafatić, and Šegvić]{grcić2023dense}
Matej Grcić, Petra Bevandić, Zoran Kalafatić, and Siniša Šegvić.
\newblock Dense out-of-distribution detection by robust learning on synthetic negative data, 2023.

\bibitem[Guo et~al.(2017)Guo, Pleiss, Sun, and Weinberger]{Guo2017ICML}
Chuan Guo, Geoff Pleiss, Yu Sun, and Kilian~Q Weinberger.
\newblock On calibration of modern neural networks.
\newblock 2017.

\bibitem[Hendrycks and Gimpel(2017)]{Hendrycks2017ICLR}
Dan Hendrycks and Kevin Gimpel.
\newblock A baseline for detecting misclassified and out-of-distribution examples in neural networks.
\newblock In \emph{ICLR}, 2017.

\bibitem[Hendrycks et~al.(2019{\natexlab{a}})Hendrycks, Mazeika, and Dietterich]{Hendrycks2019ICLR}
Dan Hendrycks, Mantas Mazeika, and Thomas Dietterich.
\newblock Deep anomaly detection with outlier exposure.
\newblock In \emph{ICLR}, 2019{\natexlab{a}}.

\bibitem[Hendrycks et~al.(2019{\natexlab{b}})Hendrycks, Mazeika, and Dietterich]{hendrycks2019oe}
Dan Hendrycks, Mantas Mazeika, and Thomas Dietterich.
\newblock Deep anomaly detection with outlier exposure.
\newblock \emph{Proceedings of the International Conference on Learning Representations}, 2019{\natexlab{b}}.

\bibitem[Hoogeboom et~al.(2019)Hoogeboom, Van Den~Berg, and Welling]{pmlr-v97-hoogeboom19a}
Emiel Hoogeboom, Rianne Van Den~Berg, and Max Welling.
\newblock Emerging convolutions for generative normalizing flows.
\newblock In \emph{Proceedings of the 36th International Conference on Machine Learning}, pages 2771--2780. PMLR, 2019.

\bibitem[Jia et~al.(2021)Jia, Yang, Xia, Chen, Parekh, Pham, Le, Sung, Li, and Duerig]{pmlr-v139-jia21b}
Chao Jia, Yinfei Yang, Ye Xia, Yi-Ting Chen, Zarana Parekh, Hieu Pham, Quoc Le, Yun-Hsuan Sung, Zhen Li, and Tom Duerig.
\newblock Scaling up visual and vision-language representation learning with noisy text supervision.
\newblock In \emph{Proceedings of the 38th International Conference on Machine Learning}, pages 4904--4916. PMLR, 2021.

\bibitem[Jiang et~al.(2018)Jiang, Kim, Guan, and Gupta]{Jiang2018NeurIPS}
Heinrich Jiang, Been Kim, Melody Guan, and Maya Gupta.
\newblock To trust or not to trust a classifier.
\newblock 2018.

\bibitem[Kirillov et~al.(2023)Kirillov, Mintun, Ravi, Mao, Rolland, Gustafson, Xiao, Whitehead, Berg, Lo, Dollar, and Girshick]{Kirillov_2023_ICCV}
Alexander Kirillov, Eric Mintun, Nikhila Ravi, Hanzi Mao, Chloe Rolland, Laura Gustafson, Tete Xiao, Spencer Whitehead, Alexander~C. Berg, Wan-Yen Lo, Piotr Dollar, and Ross Girshick.
\newblock Segment anything.
\newblock In \emph{Proceedings of the IEEE/CVF International Conference on Computer Vision (ICCV)}, pages 4015--4026, 2023.

\bibitem[Kobyzev et~al.(2021)Kobyzev, Prince, and Brubaker]{9089305}
Ivan Kobyzev, Simon~J.D. Prince, and Marcus~A. Brubaker.
\newblock Normalizing flows: An introduction and review of current methods.
\newblock \emph{IEEE Transactions on Pattern Analysis and Machine Intelligence}, 43\penalty0 (11):\penalty0 3964--3979, 2021.

\bibitem[Lakshminarayanan et~al.(2017)Lakshminarayanan, Pritzel, and Blundell]{Lakshminarayanan2017NeurIPS}
Balaji Lakshminarayanan, Alexander Pritzel, and Charles Blundell.
\newblock Simple and scalable predictive uncertainty estimation using deep ensembles.
\newblock 2017.

\bibitem[Lis et~al.(2023)Lis, Honari, Fua, and Salzmann]{10334623}
Krzysztof Lis, Sina Honari, Pascal Fua, and Mathieu Salzmann.
\newblock Detecting road obstacles by erasing them.
\newblock \emph{IEEE Transactions on Pattern Analysis and Machine Intelligence}, pages 1--11, 2023.

\bibitem[Maag et~al.(2022)Maag, Chan, Uhlemeyer, Kowol, and Gottschalk]{maag2022two}
Kira Maag, Robin Chan, Svenja Uhlemeyer, Kamil Kowol, and Hanno Gottschalk.
\newblock Two video data sets for tracking and retrieval of out of distribution objects.
\newblock In \emph{Proceedings of the Asian Conference on Computer Vision}, pages 3776--3794, 2022.

\bibitem[Minderer et~al.(2021)Minderer, Djolonga, Romijnders, Hubis, Zhai, Houlsby, Tran, and Lucic]{Minderer2021NeurIPS}
Matthias Minderer, Josip Djolonga, Rob Romijnders, Frances~Ann Hubis, Xiaohua Zhai, Neil Houlsby, Dustin Tran, and Mario Lucic.
\newblock Revisiting the calibration of modern neural networks.
\newblock 2021.

\bibitem[Mukhoti and Gal(2018)]{Mukhoti2018ARXIV}
Jishnu Mukhoti and Yarin Gal.
\newblock Evaluating bayesian deep learning methods for semantic segmentation.
\newblock 1811.12709, 2018.

\bibitem[Nalisnick et~al.(2019)Nalisnick, Matsukawa, Teh, Gorur, and Lakshminarayanan]{nalisnick2018deep}
Eric Nalisnick, Akihiro Matsukawa, Yee~Whye Teh, Dilan Gorur, and Balaji Lakshminarayanan.
\newblock Do deep generative models know what they don't know?
\newblock \emph{International Conference on Learning Representations}, 2019.

\bibitem[Nayal et~al.(2023)Nayal, Yavuz, Henriques, and Güney]{nayal2023ICCV}
Nazir Nayal, Mısra Yavuz, João~F. Henriques, and Fatma Güney.
\newblock Rba: Segmenting unknown regions rejected by all.
\newblock In \emph{ICCV}, 2023.

\bibitem[Nayal et~al.(2024)Nayal, Shoeb, and Güney]{nayal2024likelihoodratiobasedapproachsegmenting}
Nazir Nayal, Youssef Shoeb, and Fatma Güney.
\newblock A likelihood ratio-based approach to segmenting unknown objects.
\newblock \emph{arXiv preprint}, arXiv:2409.06424, 2024.

\bibitem[Nekrasov et~al.(2023)Nekrasov, Hermans, Kuhnert, and Leibe]{nekrasov2023ugains}
Alexey Nekrasov, Alexander Hermans, Lars Kuhnert, and Bastian Leibe.
\newblock {UGainS: Uncertainty Guided Anomaly Instance Segmentation}.
\newblock In \emph{GCPR}, 2023.

\bibitem[Neyman and Pearson(1933)]{neyman1933ix}
Jerzy Neyman and Egon~Sharpe Pearson.
\newblock Ix. on the problem of the most efficient tests of statistical hypotheses.
\newblock \emph{Philosophical Transactions of the Royal Society of London. Series A, Containing Papers of a Mathematical or Physical Character}, 231\penalty0 (694-706):\penalty0 289--337, 1933.

\bibitem[Nguyen et~al.(2015{\natexlab{a}})Nguyen, Yosinski, and Clune]{nguyen2015deep}
Anh Nguyen, Jason Yosinski, and Jeff Clune.
\newblock Deep neural networks are easily fooled: High confidence predictions for unrecognizable images.
\newblock In \emph{CVPR}, pages 427--436, 2015{\natexlab{a}}.

\bibitem[Nguyen et~al.(2015{\natexlab{b}})Nguyen, Yosinski, and Clune]{Nguyen2015CVPR}
Anh~M Nguyen, Jason Yosinski, and Jeff Clune.
\newblock Deep neural networks are easily fooled: High confidence predictions for unrecognizable images.
\newblock In \emph{CVPR}, 2015{\natexlab{b}}.

\bibitem[Pinggera et~al.(2015)Pinggera, Franke, and Mester]{7353537}
Peter Pinggera, Uwe Franke, and Rudolf Mester.
\newblock High-performance long range obstacle detection using stereo vision.
\newblock In \emph{2015 IEEE/RSJ International Conference on Intelligent Robots and Systems (IROS)}, pages 1308--1313, 2015.

\bibitem[Pinggera et~al.(2016)Pinggera, Ramos, Gehrig, Franke, Rother, and Mester]{lost_and_found}
Peter Pinggera, Sebastian Ramos, Stefan Gehrig, Uwe Franke, Carsten Rother, and Rudolf Mester.
\newblock Lost and found: detecting small road hazards for self-driving vehicles.
\newblock In \emph{2016 IEEE/RSJ International Conference on Intelligent Robots and Systems (IROS)}, page 1099–1106. IEEE Press, 2016.

\bibitem[Popov et~al.(2023)Popov, Gebhardt, Chen, and Oldja]{10160592}
Alexander Popov, Patrik Gebhardt, Ke Chen, and Ryan Oldja.
\newblock Nvradarnet: Real-time radar obstacle and free space detection for autonomous driving.
\newblock In \emph{2023 IEEE International Conference on Robotics and Automation (ICRA)}, pages 6958--6964, 2023.

\bibitem[Radford et~al.(2021)Radford, Kim, Hallacy, Ramesh, Goh, Agarwal, Sastry, Askell, Mishkin, Clark, Krueger, and Sutskever]{pmlr-v139-radford21a}
Alec Radford, Jong~Wook Kim, Chris Hallacy, Aditya Ramesh, Gabriel Goh, Sandhini Agarwal, Girish Sastry, Amanda Askell, Pamela Mishkin, Jack Clark, Gretchen Krueger, and Ilya Sutskever.
\newblock Learning transferable visual models from natural language supervision.
\newblock In \emph{Proceedings of the 38th International Conference on Machine Learning}, pages 8748--8763. PMLR, 2021.

\bibitem[Rai et~al.(2023)Rai, Cermelli, Fontanel, Masone, and Caputo]{Rai_2023_ICCV}
Shyam~Nandan Rai, Fabio Cermelli, Dario Fontanel, Carlo Masone, and Barbara Caputo.
\newblock Unmasking anomalies in road-scene segmentation.
\newblock In \emph{Proceedings of the IEEE/CVF International Conference on Computer Vision (ICCV)}, pages 4037--4046, 2023.

\bibitem[Rao et~al.(2022)Rao, Zhao, Chen, Tang, Zhu, Huang, Zhou, and Lu]{rao2021denseclip}
Yongming Rao, Wenliang Zhao, Guangyi Chen, Yansong Tang, Zheng Zhu, Guan Huang, Jie Zhou, and Jiwen Lu.
\newblock Denseclip: Language-guided dense prediction with context-aware prompting.
\newblock In \emph{Proceedings of the IEEE Conference on Computer Vision and Pattern Recognition (CVPR)}, 2022.

\bibitem[Reiss et~al.(2021)Reiss, Cohen, Bergman, and Hoshen]{reiss2021panda}
Tal Reiss, Niv Cohen, Liron Bergman, and Yedid Hoshen.
\newblock Panda: Adapting pretrained features for anomaly detection and segmentation.
\newblock In \emph{Proceedings of the IEEE/CVF Conference on Computer Vision and Pattern Recognition}, pages 2806--2814, 2021.

\bibitem[Roth et~al.(2022)Roth, Pemula, Zepeda, Schölkopf, Brox, and Gehler]{retrieval_industry}
Karsten Roth, Latha Pemula, Joaquin Zepeda, Bernhard Schölkopf, Thomas Brox, and Peter Gehler.
\newblock Towards total recall in industrial anomaly detection.
\newblock In \emph{2022 IEEE/CVF Conference on Computer Vision and Pattern Recognition (CVPR)}, pages 14298--14308, 2022.

\bibitem[Shoeb et~al.(2024)Shoeb, Chan, Schwalbe, Nowzad, G{\"u}ney, and Gottschalk]{shoeb2024have}
Youssef Shoeb, Robin Chan, Gesina Schwalbe, Azarm Nowzad, Fatma G{\"u}ney, and Hanno Gottschalk.
\newblock Have we ever encountered this before? retrieving out-of-distribution road obstacles from driving scenes.
\newblock In \emph{Proceedings of the IEEE/CVF Winter Conference on Applications of Computer Vision}, pages 7396--7406, 2024.

\bibitem[Sun et~al.(2022)Sun, Ming, Zhu, and Li]{sun2022knnood}
Yiyou Sun, Yifei Ming, Xiaojin Zhu, and Yixuan Li.
\newblock Out-of-distribution detection with deep nearest neighbors.
\newblock \emph{ICML}, 2022.

\bibitem[Tereshchenko and Zakala(2024)]{Tereshchenko2024}
V.M. Tereshchenko and P.A. Zakala.
\newblock Coreset discovery for machine learning problems.
\newblock \emph{Cybernetics and Systems Analysis}, 60:\penalty0 198--208, 2024.

\bibitem[Tian et~al.(2022)Tian, Liu, Pang, Liu, Chen, and Carneiro]{pebal_2022_eccv}
Yu Tian, Yuyuan Liu, Guansong Pang, Fengbei Liu, Yuanhong Chen, and Gustavo Carneiro.
\newblock Pixel-wise energy-biased abstention learning for anomaly segmentation on complex urban driving scenes.
\newblock In \emph{Computer Vision -- ECCV 2022}, pages 246--263, Cham, 2022. Springer Nature Switzerland.

\bibitem[Tokudome et~al.(2017)Tokudome, Ayukawa, Ninomiya, Enokida, and Nishida]{Tokudome2017DevelopmentOR}
Naruaki Tokudome, Shuhei Ayukawa, Shun Ninomiya, Shuichi Enokida, and Takeshi Nishida.
\newblock Development of real-time environment recognition system using lidar for autonomous driving.
\newblock In \emph{International Conference on ICT Robotics}, pages 25--26, 2017.

\bibitem[Voj{\'\i}\v{r} and Matas(2023)]{Vojir_2023_WACV}
Tom\'a\v{s} Voj{\'\i}\v{r} and Ji\v{r}{\'\i} Matas.
\newblock Image-consistent detection of road anomalies as unpredictable patches.
\newblock In \emph{Proceedings of the IEEE/CVF Winter Conference on Applications of Computer Vision (WACV)}, pages 5491--5500, 2023.

\bibitem[Vojíř et~al.(2024)Vojíř, Šochman, and Matas]{Vojir_2024_ECCV}
Tomáš Vojíř, Jan Šochman, and Jiří Matas.
\newblock {PixOOD: Pixel-Level Out-of-Distribution Detection}.
\newblock In \emph{ECCV}, 2024.

\bibitem[Williamson and Thorpe(1998)]{Williamson1998DETECTIONOS}
Todd Williamson and Charles~E. Thorpe.
\newblock Detection of small obstacles at long range using multibaseline stereo.
\newblock 1998.

\bibitem[Xu et~al.(2022)Xu, De~Mello, Liu, Byeon, Breuel, Kautz, and Wang]{9879676}
Jiarui Xu, Shalini De~Mello, Sifei Liu, Wonmin Byeon, Thomas Breuel, Jan Kautz, and Xiaolong Wang.
\newblock Groupvit: Semantic segmentation emerges from text supervision.
\newblock In \emph{2022 IEEE/CVF Conference on Computer Vision and Pattern Recognition (CVPR)}, pages 18113--18123, 2022.

\bibitem[Zhang and Wischik(2022)]{zhang2022falsehoods}
Andi Zhang and Damon Wischik.
\newblock Falsehoods that ml researchers believe about ood detection.
\newblock In \emph{NeurIPS ML Safety Workshop}, 2022.

\bibitem[Zheng~Ding(2023)]{ding2023maskclip}
Zhuowen~Tu Zheng~Ding, Jieke~Wang.
\newblock Open-vocabulary universal image segmentation with maskclip.
\newblock In \emph{International Conference on Machine Learning}, 2023.

\bibitem[Zou et~al.(2023)Zou, Yang, Zhang, Li, Li, Wang, Wang, Gao, and Lee]{zou2023segment}
Xueyan Zou, Jianwei Yang, Hao Zhang, Feng Li, Linjie Li, Jianfeng Wang, Lijuan Wang, Jianfeng Gao, and Yong~Jae Lee.
\newblock Segment everything everywhere all at once.
\newblock In \emph{Advances in Neural Information Processing Systems}, 2023.

\bibitem[Zou et~al.(2022)Zou, Jeong, Pemula, Zhang, and Dabeer]{Zou_22}
Yang Zou, Jongheon Jeong, Latha Pemula, Dongqing Zhang, and Onkar Dabeer.
\newblock Spot-the-difference self-supervised pre-training for anomaly detection and segmentation.
\newblock In \emph{Computer Vision -- ECCV 2022}, pages 392--408, Cham, 2022. Springer Nature Switzerland.

\end{thebibliography}
}


\end{document}